\documentclass[10pt,journal,compsoc]{IEEEtran}
\ifCLASSOPTIONcompsoc
  \usepackage[nocompress]{cite}
\else
  \usepackage{cite}
\fi
\ifCLASSINFOpdf
\else
\fi
\usepackage{amsmath}
\usepackage{bm}
\usepackage{bbm}
\usepackage{amssymb}
\usepackage{amsfonts}
\usepackage{mathtools}

\usepackage{url}
\newcommand\MYhyperrefoptions{
    bookmarks=true,
    bookmarksnumbered=true,
    pdfpagemode={UseOutlines},
    plainpages=false,
    pdfpagelabels=true,
    colorlinks=true,
    pdftitle={Improving Prediction Performance and Model Interpretability through Attention Mechanisms from Basic and Applied Research Perspectives},%<!CHANGE!
    pdfsubject={Ph.D. thesis},%<!CHANGE!
    pdfauthor={Shunsuke Kitada},%<!CHANGE!
    pdfkeywords={Natural Language Processing, Computational Advertising, Attention Mechanism, Adversarial Training, Virtual Adversarial Training, Interpretability, Ad Conversion Prediction, Ad Discontinuation Prediction}
}%<^!CHANGE!
\ifCLASSINFOpdf
    \usepackage[\MYhyperrefoptions,pdftex]{hyperref}
\else
    \usepackage[\MYhyperrefoptions,breaklinks=true,dvips]{hyperref}
    \usepackage{breakurl}
\fi
\begin{document}
\bstctlcite{IEEEexample:BSTcontrol}

\title{
    Improving Prediction Performance and Model \\ 
    Interpretability through Attention Mechanisms \\ 
    from Basic and Applied Research Perspectives
}

\author{
    Shunsuke Kitada \\
    Applied Informatics, Graduate School of Science and Engineering, Hosei University, Tokyo, Japan \\
    Supervisor: Hitoshi Iyatomi
}


\markboth{The bulletin of Graduate School of Science and Engineering, Hosei University, Vol.64 (03/2023)}%
{Shunsuke Kitada: Improving Prediction Performance and Model Interpretability through Attention Mechanisms from Basic and Applied Research Perspectives}



\IEEEtitleabstractindextext{%
\begin{abstract}
With the dramatic advances in deep learning technology, machine learning research is focusing on improving the interpretability of model predictions as well as prediction performance in both basic and applied research. 
While deep learning models have much higher prediction performance than traditional machine learning models, the specific prediction process is still difficult to interpret and/or explain. 
This is known as the black-boxing of machine learning models and is recognized as a particularly important problem in a wide range of research fields, including manufacturing, commerce, robotics, and other industries where the use of such technology has become commonplace, as well as the medical field, where mistakes are not tolerated.
This bulletin is based on the summary of the author's dissertation.
The research summarized in the dissertation focuses on the attention mechanism, which has been the focus of much attention in recent years, and discusses its potential for both basic research in terms of improving prediction performance and interpretability, and applied research in terms of evaluating it for real-world applications using large data sets beyond the laboratory environment.
The dissertation also concludes with a summary of the implications of these findings for subsequent research and future prospects in the field. 
\end{abstract}

}

\maketitle

% is not selected <OR> if conference mode is selected - because compsoc
% conference papers position the abstract like regular (non-compsoc)
\IEEEdisplaynontitleabstractindextext

% \fi
\IEEEpeerreviewmaketitle

\ifCLASSOPTIONcompsoc
\IEEEraisesectionheading{\section{Introduction}\label{sec:introduction}}
\else
\section{Introduction}
\label{sec:introduction}
\fi


\IEEEPARstart{D}{eep} learning~\cite{lecun2015deep} is one of the machine learning (ML) models and has contributed greatly to the current development of artificial intelligence (AI).
Compared to traditional ML models such as decision trees~\cite{breiman1984classification} and support vector machines~\cite{hearst1998support}, DL models have been shown to dramatically improve prediction performance in various fields because they can automatically learn important features from data to realize a goal through the training.
DL models, which do not require knowledge about the subject, have demonstrated predictive and generative abilities beyond human capabilities, especially in the fields of computer vision (CV)~\cite{krizhevsky2012imagenet, he2016deep} and natural language processing (NLP)~\cite{vaswani2017attention, devlin2019bert}.
Since DL models will be used more frequently in the future, it is necessary for users to be able to interpret the validity of the prediction results of these models and the basis for them, in terms of the reliability and practicality of the models.

While the recent DL model has excellent prediction performance, it is difficult to interpret and explain the model prediction due to the complex architecture of the model. 
This black box and/or not transparent nature is an important issue that needs to be resolved~\cite{castelvecchi2016can}.
Explainable AI (XAI) is a field that seeks to explain the predictions of ML/DL models.
This field has been studied for more than 40 years~\cite{scott1977explanation, swartout1985explaining}, and classical XAI provided explanations based on roles constructed by humans carefully.
Recent DL models have complex neural network (NN) structures consisting of various nonlinear transformations, making a human interpretation of the internal inference process very difficult.
The interpretability of the predictions is recognized as a particularly important issue in a wide range of research fields, including manufacturing~\cite{sharp2018survey}, e-commerce~\cite{zhang2020explainable}, and robotics~\cite{karoly2020deep}, where the use of DL models is becoming common, as well as in the medical field~\cite{singh2020explainable} and autonomous driving~\cite{omeiza2021explanations}, where mistakes are not tolerated.

\subsection{Objectives and Requirements for XAI}

Explainability for ML/DL models can be taken to mean a variety of things.
In this section, we clarify our position on the interpretation and usage of these terms for ML/DL models in this work, with examples of what is stated in the literature in the related fields.
Previous studies have attempted to clarify the purpose and requirements for XAI~\cite{arrieta2020explainable, adadi2018peeking, guidotti2018survey}.
Additionally, there are examples of companies such as Google~\footnote{\url{https://cerre.eu/wp-content/uploads/2020/07/ai_explainability_whitepaper_google.pdf}} and Amazon~\footnote{\url{https://pages.awscloud.com/rs/112-TZM-766/images/Amazon.AI.Fairness.and.Explainability.Whitepaper.pdf}}, which place ML at the center of their business, discussing the provision of explanations from the standpoint of providing their systems and services.

XAI should consider the audience because the contents and details of the reasons to be presented by depending on the audience.
As a promising definition of XAI, Arrieta et al.~\cite{arrieta2020explainable} define it as follows:
\begin{quote}
    \textit{Given an audience, an explainable Artificial Intelligence is one that produces details or reasons to make its functioning clear or easy to understand.}
\end{quote}
Additionally, Arrieta et al.~\cite{arrieta2020explainable}, in light of the previous XAI research, have identified the following nine goals for the research: (1) \textbf{trustworthiness}, (2) \textbf{causality}, (3) \textbf{transferability}, (4) \textbf{informativeness}, (5) \textbf{confidence}, (6) \textbf{fairness}, (7) \textbf{accessibility}, (8) \textbf{interactivity}, (9) \textbf{privacy awareness}.

We agree with the above definition of Arrieta et al.~\cite{arrieta2020explainable} and aim to achieve the following goal for the audiences: ``\textit{building XAI that provides reasons and details that make understanding easier.}''
As described below, this work is divided into discussions of basic and applied perspectives, each of which aims to define and improve ``model interpretability'' as follows.
In the basic research perspective, the goal is to provide an interpretation that makes the basis for the predictions of multiple interpretation methods identical.
This goal focuses on (1) trustworthiness, which is the state of trust that the model will function as intended when the problem is solved, and (5) confidence, which is the state of stability in the behavior of the model.
In the applied research perspective, assuming that the audience is the operator of the service, the goal is to provide an interpretation that better supports the operator's decision-making.
This goal focuses on (4) informativeness, the state in which humans can extract the information necessary for decision-making when solving real-world problems.

Throughout this work, we will focus on the interpretation of each word in the input sentence/document in the prediction of NLP models.
By making it possible for audiences to interpret where the model contributes predictions to the input, we believe that the above definition of XAI is satisfied.
The ability to interpret the contribution of inputs from the DL model is useful for validating the model behavior, analyzing errors, and making decisions when operating the model in the real world.

\subsection{XAI Methods to Interpret the Prediction Results}

There are two major approaches to the black box nature of DL models: (1) designing transparent models~\cite{lundberg2020local,saarela2021comparison,banerjee2006convex} and (2) providing post-hoc explanations for model predictions~\cite{selvaraju2017grad,sundararajan2017axiomatic,zhang2020explainable}.
For (1), the design of transparent models involves research into understanding the architecture of the model itself~\cite{guidotti2018survey} and the learning algorithms of the model\cite{banerjee2006convex}.
In DL models, these works have limited applicability to the target model architecture, making them difficult to apply to existing models. 
They typically have limited prediction performance as they attempt to provide human-interpretable explanations.
For (2), the post-hoc explanation methods involve visualization of factors that influence predictions~\cite{simonyan2013deep, sundararajan2017axiomatic}, and providing analytical explanations with concrete examples if applicable~\cite{kenny2021explaining}.
The post-hoc approach is widely used today because it generally applies to DL models and is easier than designing transparent models.

In the following, we explain the visualization approaches of importance based on the gradient for the prediction result and the learned attention weights in the post-hoc explanation, which is nowadays the mainstream interpretation of the predictions.
First, we define the formulas that will be used throughout this work with respect to these approaches.

We consider a recurrent-neural network (RNN) model for NLP task.
The input of the model is word sequence $\bm{x} = (x_1, x_2, \cdots, x_T) \in \mathcal{V}^T$ of the length $T$ where the words taken from vocabulary $\mathcal{V}$.
The output of the model is $\hat{\bm{y}} \in \mathbb{R}^{\mathcal{C}}$ corresponds to ground truth $\bm{y} \in \mathbb{R}^{\mathcal{C}}$, where $\mathcal{C}$ is the set of class labels.
We introduce the following short notation for the word sequence $(x_1, x_2, \cdots, x_T)$ as $(x_t)_{t=1}^{T}$.
Let $\bm{w}_t \in \mathbb{R}^{d}$ be a $d$-dimensional word embedding corresponding to $x_t$.
We represent each word with the word embeddings to obtain $(\bm{w}_t)_{t=1}^{T} \in \mathbb{R}^{d \times T}$.
The word embeddings is encoded with encoder $\textbf{Enc}$ to obtain the $m$-dimensional hidden states:
\begin{equation}
    \bm{h}_t = \textbf{Enc}(\bm{w}_t, \bm{h}_{t-1}) \in \mathbb{R}^{m},
\end{equation}
where $\bm{h}_0$ is the initial hidden state, and it is regarded as a zero vector.

\subsubsection{Gradient-based Approach}

The gradient-based approach estimates the contribution of input $\bm{x}$ to ground truth $\bm{y}$ or prediction $\hat{\bm{y}}$ by computing the partial derivative of $\bm{x}$ with respect to $\bm{y}$ or $\hat{\bm{y}}$. 
Here we use the term gradient for $\partial \bm{y} / \partial \bm{x}$.
The goal of the gradient-based approach is to estimate attribution maps $\bm{g}^c = (g_i^c)_{i=1}^{T}$ for each word.
The attribution maps $\bm{g}^c$ are considered to capture the importance of each input word for a particular output class $c \in \mathcal{C}$.

Following the advent of AlexNet~\cite{krizhevsky2012imagenet}, shortly after DL first came into the limelight, an early gradient-based approach was proposed by Simonyan et al.~\cite{simonyan2013deep} and has long supported the interpretation of DL model predictions.
They used a formulation similar to the above to present the attribute map to the user as a saliency map based on its absolute value:
\begin{equation}
    \bm{g}^{c}_{\mathrm{saliency}} = \left|\frac{\partial \bm{y}}{\partial \bm{x}}\right|.
\end{equation}
The core idea of gradient-based methods is to map gradient information \textit{backward} into the input space based on labels or inference results.

Because gradient-based approaches are applicable to all DL models trained by backpropagation, they have been used for many years in various fields for insight into the internal workings of models and for error analysis~\cite{selvaraju2017grad, shrikumar2017learning, sundararajan2017axiomatic}.
Since then, gradient-based approaches such as GradCAM~\cite{selvaraju2017grad}, DeepLIFT~\cite{shrikumar2017learning}, and Integrated Gradient~\cite{sundararajan2017axiomatic} have been proposed to provide interpretations of predictions that provide deeper human insight, different from the rule-based approaches that were common during classical XAI.
There is a large body of literature claiming that gradient-based approaches can be used to explain the importance of input features~\cite{aubakirova2016interpreting,karlekar2018detecting}.

Gradient-based approaches have a variety of advantages.
First, gradient-based methods are fast and efficient in the computation of attribution maps.
It is easily scalable because it does not depend on the number of input features, and it can be easily computed with high performance with the support of deep learning frameworks.
Second, it can be applied to existing models and any network architecture with very few lines of code by overriding the gradient of nonlinearity in the computational graph.
It is easy to implement because there is no need to implement custom layers or operations.

On the other hand, there are limitations to gradient-based approaches.
The biggest problem is that the attribution maps presented are often visually noisy~\cite{smilkov2017smoothgrad}.
The gradient-based approaches assume that small changes in input features cause small changes in predictions, which may not always be the case~\cite{koh2017understanding}.
Additionally, they are sensitive to the choice of input baseline, which can affect the attribution scores~\cite{sundararajan2017axiomatic, ancona2017towards}.
The gradient-based approaches are not able to capture interactions between features and, therefore, may not provide a complete explanation of the model's predictions~\cite{fong2017interpretable}.

\subsubsection{Attention-based Approach}

In recent DLs, attention mechanisms~\cite{bahdanau2014neural,vaswani2017attention} are used to focus more \textit{attention} on specific parts of the input as the model processes it.
The effectiveness of the mechanism was initially validated in machine translation~\cite{bahdanau2014neural} and image captioning~\cite{xu2015show}, but it is now being expanded in a wide variety of tasks.
The attention mechanism works by assigning weights to each part of the input.

The key component in the attention mechanism is an attention score function $\mathcal{S}(\cdot, \cdot)$.
The score function maps a query $\bm{Q}$ and key $\bm{K}$ to attention score $\tilde{a}_t$ for the $t$-th word.
For the NLP tasks, we consider $\bm{K} = (\bm{h}_t)_{t=1}^{T}$.
The attention scores $\tilde{\bm{a}} = (\tilde{a}_t)_{t=1}^{T}$ projected to sum to 1 by a alignment function $\mathcal{A}$, which results in the attention weights $\bm{a}$:
\begin{equation}
    \bm{a} = \mathcal{A}(\bm{K}, \bm{Q}) = \phi(\mathcal{S}(\bm{K}, \bm{Q})),
\end{equation}
where $\phi$ is a projection function to probability, such as softmax.
The weight $\bm{a}$ indicates the importance of that portion to the current task.
These weights are then used to selectively focus on the most important parts of the input to generate output.
We then compute weighted instance vector $\bm{h}_{\bm{a}}$ as a weighted sum of the hidden states:
\begin{equation}
    \bm{h}_{\bm{a}} = \sum_{i=1}^{T} a_{t} \bm{h}_t
\end{equation}
Finally, $\bm{h}_{a}$ is fed to a prediction layer, which outputs a probability distribution over the set of categories $\mathcal{C}$.

The score function conventionally uses the following additive attention formulation:
\begin{equation}
    \mathcal{S}_{\mathrm{add}}(\bm{K}, \bm{Q})_i = \bm{v}^\top \tanh{(\bm{W}_1 \bm{K}_t + \bm{W}_2 \bm{Q})},
\end{equation}
where $W_1, W_2 \in \mathbb{R}^{d \times M}$ and $\bm{v} \in \mathbb{R}^{d}$ are the learnable parameters.
Currently, the Transformer and other variants often use the following scaled-dot product attention formulation:
\begin{equation}
    \mathcal{S}_{\mathrm{prod}}(\bm{K}, \bm{Q}) = \frac{\bm{K}^\top \bm{Q}}{\sqrt{m}}
\end{equation}
In this summary, each evaluation is carried out with additive attention formulation as the score function.

While the gradient-based approach can interpret the prediction of a model without any interpretation/explanation mechanism, the attention-based DL model allows for the interpretation of the model by visualizing the learned attention weights.
Explanations based on learned weights of attention mechanisms~\cite{bahdanau2014neural} can be shown in a \textit{forward} direction for DL model inputs, giving a clear explanation that appeals to human intuition.
Since the attention mechanism introduced in conventional RNN is generally placed before the last layer (or the last layer of the encoder in encoder-decoder architecture), the attention weights are learned to emphasize the part that contributes to the final prediction.
For the Transformer model~\cite{vaswani2017attention}, which has been attracting much interest in recent years, attention rollout~\cite{abnar2020quantifying} has been proposed, which visualizes the result of the matrix product of the attention scores.

\subsection{Discussion on Post-hoc Explanation Approaches}

Although the attention mechanism contributes significantly to predictive performance and model interpretability, researchers find that the mechanism still suffers in the predictive interpretation of the model.
In a claim that surprised the field, Jain and Wallace~\cite{jain2019attention} argued that ``attention is not an explanation.''
They reported the following two major problems in a simple RNN-based model with the attention mechanism, especially by NLP task:
(1) There is not necessarily a strong correlation between the regions estimated to be important by attention weights and those obtained by gradients.
(2) Small perturbations to the attention mechanism lead to unintended predictive changes, and those adversarial perturbations that deceive the mechanism lead to large predictive errors.
For (1), the authors reported that Kendall's rank correlations~\cite{kendall1938new} of importance indicated by the learned attention weight and word importance calculated by the gradient show almost no correlation.
For (2), they found that small perturbations to the attention mechanisms lead to unintended prediction changes, while adversarial perturbations that deceive the mechanism lead to large prediction errors.

Against the weak explanatory nature of the attention mechanisms reported above, refutational analyses and methods to overcome them have been proposed in recent years~\cite{wiegreffe2019attention,serrano2019attention,grimsley2020attention,wang2016attention}.
On the analytical side, Wiegreffe and Pinter~\cite{wiegreffe2019attention} argued that Jain and Wallace~\cite{jain2019attention}'s claim depends on the definition of explanation and that testing it requires considering different aspects of the model with attention mechanisms in a rigorous experimental setting.
Furthermore, similar to Wiegreffe and Pinter~\cite{wiegreffe2019attention}, several studies argue that the effects of explanations by attention mechanisms vary depending on the definition of explanatory properties~\cite{serrano2019attention, grimsley2020attention, wang2020gradient}.
With regard to improving the interpretability of the attention mechanism, there are some studies that re-examined the structure of the mechanism itself~\cite{mrini2020rethinking} or proposed a training technique that makes the attention mechanism robust to perturbations \textit{indirectly}~\cite{bento2021timeshap}.
On the other hand, to our knowledge, no technique has been proposed to \textit{directly} address the vulnerability of attention mechanisms to perturbation, as pointed out in Jain and Wallace~\cite{jain2019attention}.

In the light of the definition of XAI in the Arrieta et al.~\cite{arrieta2020explainable}, if each interpretation approach provides a different interpretation, it will have a negative impact on the audience, especially in the trustworthiness and confidence.
As described above, various studies have been conducted on the interpretability of DL models, focusing on the gradient- and attention-based approaches. 
However, a comprehensive synthesis of these multiple approaches to obtain reliable and robust interpretations is an important basic research direction, and there remains room for further research.

\subsection{Interpreting Deep Learning Models in Real-World Applications}

While the above discussion focused on the interpretation of DL models in the basic research perspectives, this section describes the provision of interpretation in DL models that are operated in the real world.
To begin with, we emphasize that there are still few studies that have validated using real-world datasets under a practical situation.
We speculate that this may be because information about applications in the real world is often confidential.
Therefore, DL-based models, including attention-based models, have been developed in fields such as machine translation~\cite{bahdanau2014neural, vaswani2017attention}, machine reading comprehension~\cite{devlin2019bert}, and image classification~\cite{dosovitskiy2020image}.
However, the development and evaluation of such models have been carried out on relatively small and well-organized data sets, limited to so-called laboratory environments.
One of the root causes of the limitation is the lack of publicly available dataset, which is important for the research area.
The effectiveness of DL models outside of these data-available areas remains to be developed.
So far, attention-based DL models have been reported to perform well on various tasks.
On the other hand, it remains unclear how the DL models perform on noisier, more imbalanced, and diverse real-world data that may deviate from the benchmark dataset.

The literature evaluating its interpretability in DL models operating in the real world is even less extensive than the literature evaluating its performance on real-world datasets.
According to Arrieta et al.~\cite{arrieta2020explainable} and the industry tech giants, XAI is expected to be an AI that provides such audience/operators with details and reasons to clarify their own behavior or to facilitate understanding.
Furthermore, there is a certain need to build a DL model that can provide high prediction performance and prediction evidence, which have not been achieved by conventional ML models.
In sum, there is currently limited research on developing models that can be interpreted for practical use of large real-world data.
This point remains an important applied research issue.

\subsection{The Structure of the Dissertation}

The remainder of the dissertation on which this bulletin is based is organized as follows.
The first two are chapters on aspects of basic research, and the next two are chapters on aspects of applied research. 

\begin{itemize}
    \item \textbf{Chapter 2} describes adversarial training for attention mechanisms based on~\cite{kitada2021attention}.
    \item \textbf{Chapter 3} describes virtual adversarial training for attention mechanisms based on~\cite{kitada2022making}.
    \item \textbf{Chapter 4} describes conversion prediction for ad creatives based on~\cite{kitada2019conversion}.
    \item \textbf{Chapter 5} describes discontinuation prediction for ad creatives based on~\cite{kitada2022ad}.
    \item \textbf{Chapter 6} provides a discussion of the applicability, interpretability, and development of the proposed method throughout the dissertation.
    \item \textbf{Chapter 7} contains the conclusion section of the dissertation.
\end{itemize}

% for IEEE Computer Society journal papers produced under \LaTeX\ using
% IEEEtran.cls version 1.8b and later.
% % You must have at least 2 lines in the paragraph with the drop letter
% % (should never be an issue)
% I wish you the best of success.

 

\section{Basic Research Perspectives}
    In terms of the basic research for this work, we describe the vulnerability of the attention mechanism, which is essential to DL models, to perturbations and countermeasures against it.
We further discuss the interpretability of our technique after its application.
The majority of prior studies have suggested that interpretation is possible by investigating where the attention mechanisms assign large weight to the model inputs.
On the other hand, we believed that the mechanism was vulnerable to noise, which would negatively affect prediction performance and interpretability.

To overcome the above challenges, a natural idea is: to introduce adversarial perturbation to the attention mechanism that is vulnerable to perturbation so that it learns to be robust to noise.
We focus on adversarial training (AT)~\cite{goodfellow2014explaining} to deal with adversarial examples~\cite{szegedy2013intriguing} that produce inaccurate model output.
AT~\cite{goodfellow2014explaining} was first proposed in the field of image recognition as a method to overcome the weak point where the model can be fooled by input small noise/perturbation that is imperceptible to humans.
Since then, AT has been introduced in the NLP field and has demonstrated its usefulness in areas which is difficult for the DL model to predict, even when the text is mixed with metaphorical expressions that are understandable to humans (e.g., fake news detection~\cite{tariq2022adversarial}).
While AT in NLP is often applied to the input space, its effects when applied to attention mechanisms remain unclear.

The basic research perspective of this work proposes a new training technique that focuses on the vulnerability to perturbations of the attention mechanism and contributes significantly to prediction performance and prediction interpretation.
We consider using AT, in which perturbations are applied to deceive the mechanisms, exploiting adversarial perturbations.
By employing the proposed technique, DL model will be trained to pay stronger attention to areas that are more important for prediction, which is expected to not only improve prediction performance but also improve model interpretability.

\begin{figure}[t]
    \centering
    \includegraphics[width=\linewidth]{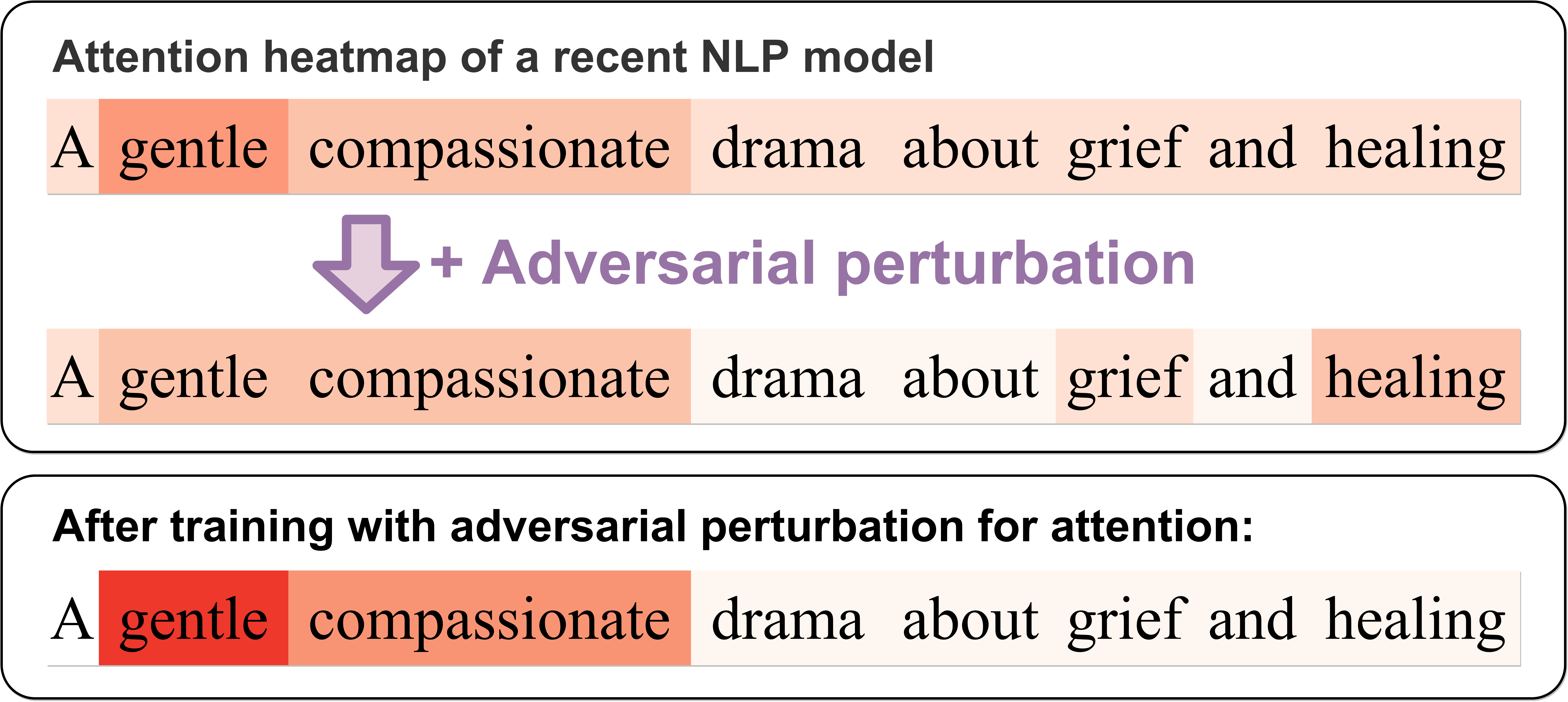}
    \caption[An example of an attention heatmap for a BiRNN model with attention mechanisms and the model with attention mechanisms trained with adversarial training]{An example of an attention heatmap for a BiRNN model with attention mechanisms and the model with attention mechanisms trained with adversarial training from the Stanford Sentiment Treebank (SST)~\cite{socher2013recursive}. The proposed adversarial training for attention mechanisms helps the model learn cleaner attention.}
    \label{fig:kitada2021attention/figure1}
\end{figure}

\subsection{AT for Attention Mechanism}\label{sec:at_for_attention}

Inspired by AT~\cite{goodfellow2014explaining}, a powerful regularization technique for enhancing model robustness, we aim to overcome the vulnerability of the attention mechanism to perturbations.
We propose a general training technique for natural language processing tasks, including AT for attention (Attention AT) and more interpretable AT for attention (Attention iAT) as shown in Fig.~\ref{fig:kitada2021attention/figure1}. 
The proposed techniques improved the prediction performance and the model interpretability by exploiting the mechanisms with AT. 
In particular, Attention iAT boosts those advantages by introducing adversarial perturbation, which enhances the difference in the attention of the sentences.
Evaluation experiments with ten open datasets revealed that AT for attention mechanisms, especially Attention iAT, demonstrated (1) the best performance in nine out of ten tasks and (2) more interpretable attention (i.e., the resulting attention correlated more strongly with gradient-based word importance) for all tasks.
Additionally, the proposed techniques are (3) much less dependent on perturbation size in AT.

\begin{figure}[t]
    \centering
    \includegraphics[width=0.8\linewidth]{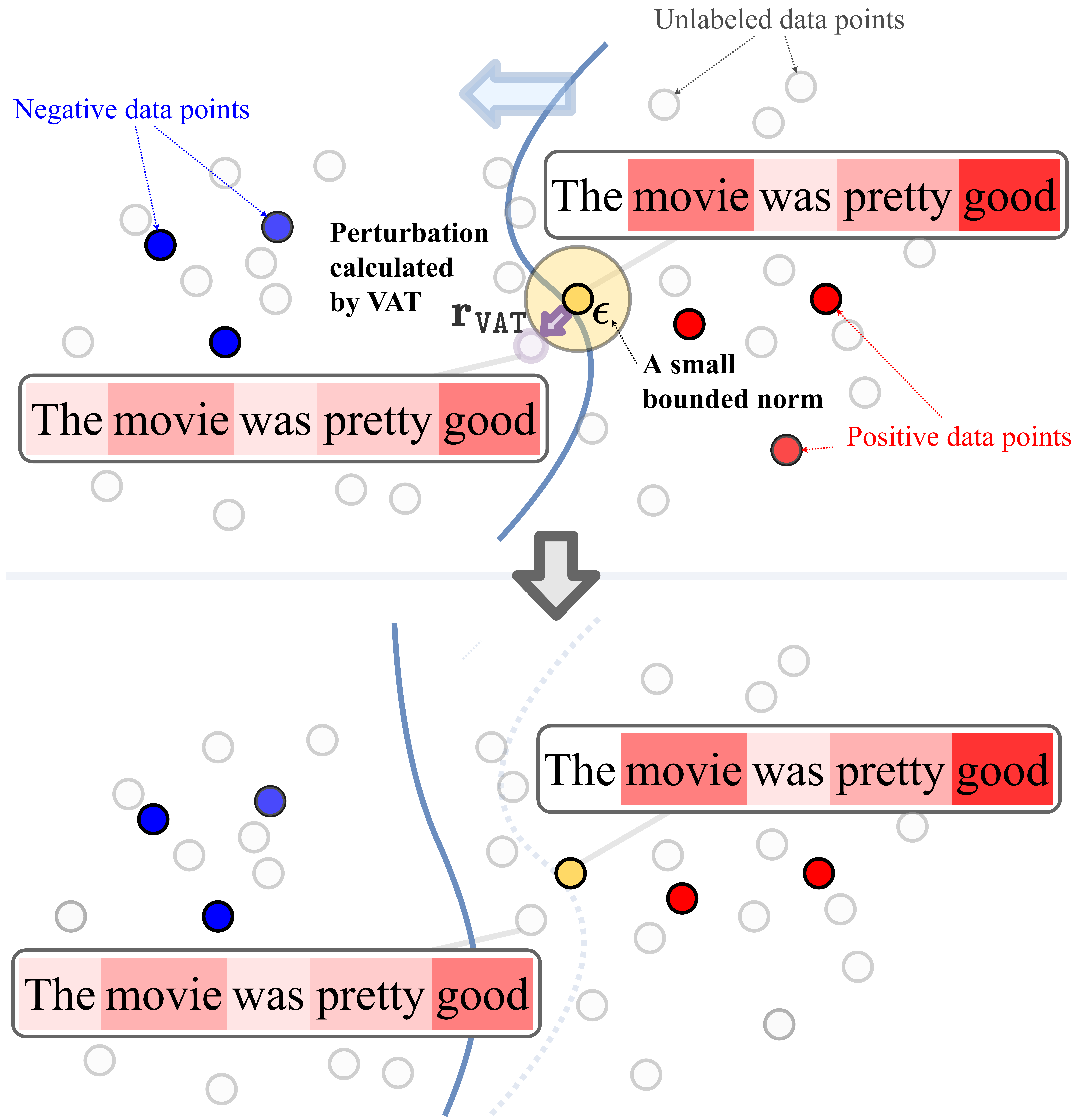}
    \caption{
        Intuitive illustration of the proposed VAT for attention mechanisms.
        Our technique can learn clearer attention by overcoming adversarial perturbations $\bm{r}_{\texttt{VAT}}$, thereby improving model interpretability
    }
    \label{fig:kitada2022making/figure1}
\end{figure}

\subsection{VAT for Attention Mechanism}\label{sec:vat_for_attention}

AT has successfully reduced the disadvantage of being vulnerable to perturbations to the attention mechanisms by considering adversarial perturbations. 
However, this technique requires label information, and thus, its use is limited to supervised settings. 
We explore the approach of incorporating virtual AT (VAT)~\cite{miyato2018virtual} into the attention mechanisms, by which adversarial perturbations can be computed even from unlabeled data.
To realize this approach, we propose two general training techniques, namely VAT for attention mechanisms (Attention VAT) and ``interpretable'' VAT for attention mechanisms (Attention iVAT), which extend AT for attention mechanisms to a semi-supervised setting, as shown in Fig.~\ref{fig:kitada2022making/figure1}.
In particular, Attention iVAT focuses on the differences in attention; thus, it can efficiently learn clearer attention and improve model interpretability, even with unlabeled data. 
Empirical experiments based on six public datasets revealed that our techniques provide better prediction performance than conventional AT-based as well as VAT-based techniques, and stronger agreement with evidence that is provided by humans in detecting important words in sentences. 
Moreover, our proposal offers these advantages without needing to add the careful selection of unlabeled data. 
That is, even if the model using our VAT-based technique is trained on unlabeled data from a source other than the target task, both the prediction performance and model interpretability can be improved.

\section{Applied Research Perspectives}
    In terms of applied research for this work, we describe research on applications of ML/DL models in NLP tasks.
In particular, we focus on applications to the field of computational advertising~\cite{dave2014computational}, which has a significant impact on business.
The field of computational advertising is a relatively new and important business-related research topic, dealing with very large online data volumes of the order of 100 million.
Display advertising is a type of online advertising in which an advertiser pays a publisher to display graphical material on the publisher's web page or application.
This graphical material, which primarily contains images and text, is commonly referred to as ad creative and serves to effectively provide product information to consumers who are willing to buy.
The market for digital advertising has been expanding enormously and is expected to grow further in the future.
In fact, the IAB internet advertising revenue report 2021~\cite{iab2021internet} states that ``digital advertising revenue increased 35.4\% year over year, the highest growth since 2006.''
Since it is difficult to operate ad data due to its large scale manually, a methodology is expected to provide operational support by means of a computer to distribute highly effective ads and discontinue ineffective ads.

Mainstream research in online advertising estimates click-through rate (CTR) and conversion rates (CVR) for users of the ads being served.
For such tasks, various methods based on ML/DL have recently emerged~\cite{chakrabarti2008contextual,richardson2007predicting,chen2016deep,covington2016deep,cheng2016wide}, including the variants of factorization machines~\cite{rendle2010factorization,juan2016field,juan2017field,guo2017deepfm} that can consider feature interactions.
While these have been evaluated and assessed on anonymized ad-serving benchmark datasets such as Criteo\footnote{\url{http://labs.criteo.com/2013/12/download-terabyte-click-logs/}} and Avazu\footnote{\url{https://www.kaggle.com/c/avazu-ctr-prediction}} and have reported some improvement in prediction performance, very few studies have evaluated them on other data, including real-world data~\cite{mishra2019guiding,zhou2020recommending}.
This is because ad data is generally very complex in terms of rights involving a large number of stakeholders, making it difficult to disclose.
Additionally, there are very few studies on the prediction of effectiveness for ads with no delivery performance, the so-called cold start setting~\cite{gope2017survey}.

The applied aspect of this work describes the practical application of NLP technology to the computational advertising field and discusses the prediction performance and model interpretability of the proposed frameworks.
We describe a framework for evaluating ad creatives that are more effective in terms of serving.
Although it is important to evaluate in advance what kind of ad creatives should be served, it is also important to provide evidence as to why the ad creatives are effective in order to improve newly created ad creatives and judge the validity of such evaluations.
We also describe a framework for automating the discontinuation of ad creatives with less serving effectiveness.
Predicting the timing of ad discontinuation itself is important.
Additionally, providing some evidence as to why the ad creative is discontinued at a certain time is an important indicator for advertisers and ad operators to consider whether to trust the prediction.
This is a major factor in confidence in the DL model.

\begin{figure}[t]
    \centering
    \includegraphics[width=\linewidth]{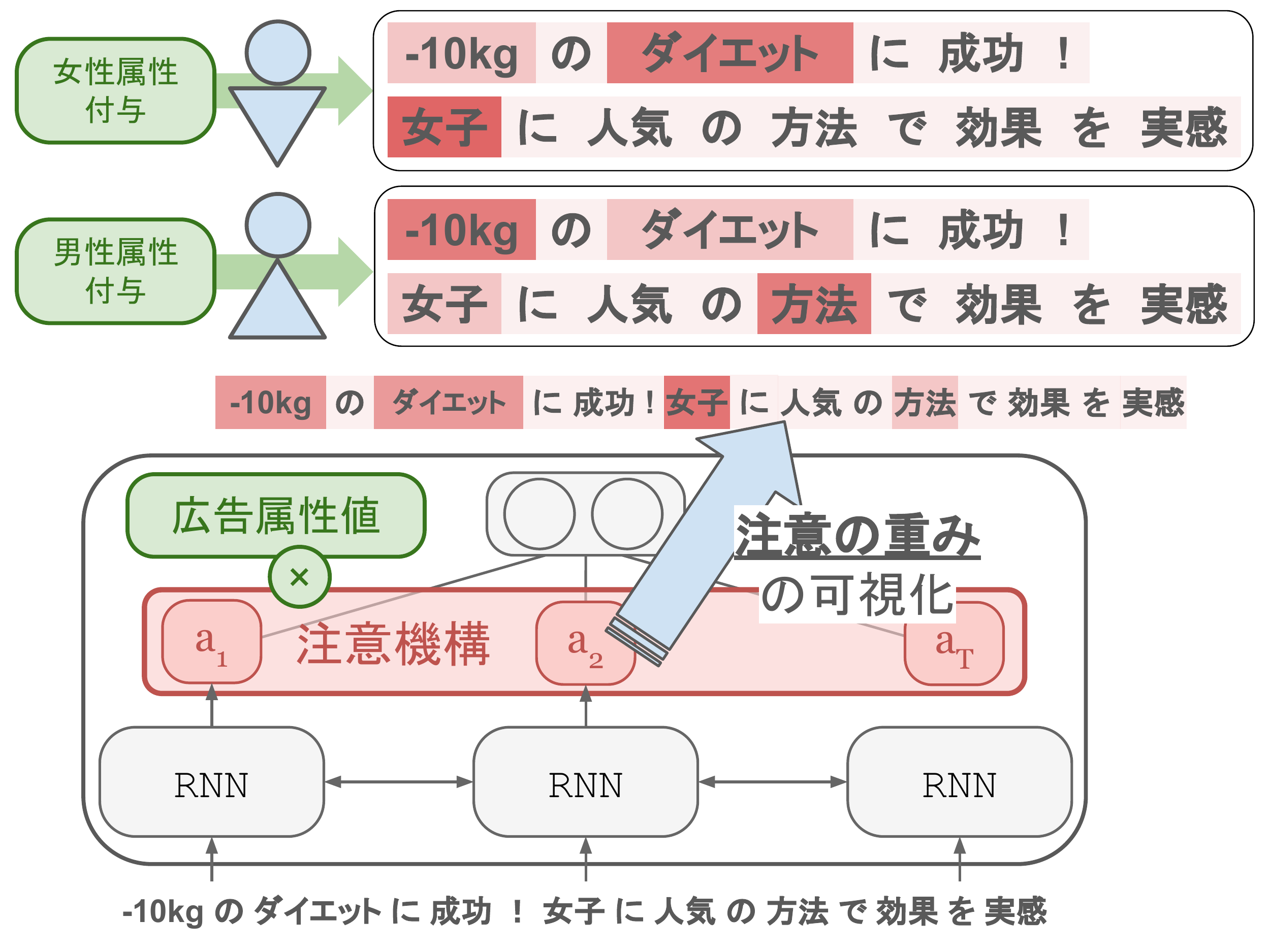}
    \caption[Outline of the proposed framework]{
        Outline of the proposed framework.
        In the framework, we propose two strategies: multi-task learning, which simultaneously predicts conversions and clicks, and a conditional attention mechanism, which detects important representations in ad creative text according to the text's attributes.
    }
    \label{fig:kitada2019conversion/proposed_architecture}
\end{figure}

\subsection{Operational Support for More Effective Ad}

Accurately predicting conversions in advertisements is generally a challenging task because such conversions do not occur frequently.
We propose a new framework to support creating high-performing ad creatives, including the accurate prediction of ad creative text conversions before serving them to the consumer, as shown in Fig.~\ref{fig:kitada2019conversion/proposed_architecture}.
The proposed framework includes the key ideas needed to train the model: multi-task learning and conditional attention mechanism.
The multi-task learning is an idea to improve prediction performance by simultaneously predicting clicks and conversions in order to overcome the difficulty of prediction due to data imbalance.
The conditional attention mechanism is a new mechanism that can take into account the attribute values of the ad creative, such as the genre of the ad creative and the gender of the delivery target, to further improve the performance of conversion prediction.
We evaluated the proposed framework on actual large-scale serving history data and confirmed that these ideas improved the performance of conversion prediction.

The conditional attention mechanism incorporated in the proposed framework is capable of interpreting word expressions that contribute to the effectiveness of ad serving.
Attention highlighting using the learned conditional attention mechanism predicts conversions in advance, taking into account the attribute values set at the time of ad submission, and simultaneously visualizes the importance of the words that contributed to the prediction.
By observing the attention highlighting when the proposed method predicts a high number of conversions for a prototype ad creative, it is possible to confirm the word expressions that better fit the target ad attributes.

\begin{figure*}[t]
    \centering
    \includegraphics[width=0.95\linewidth]{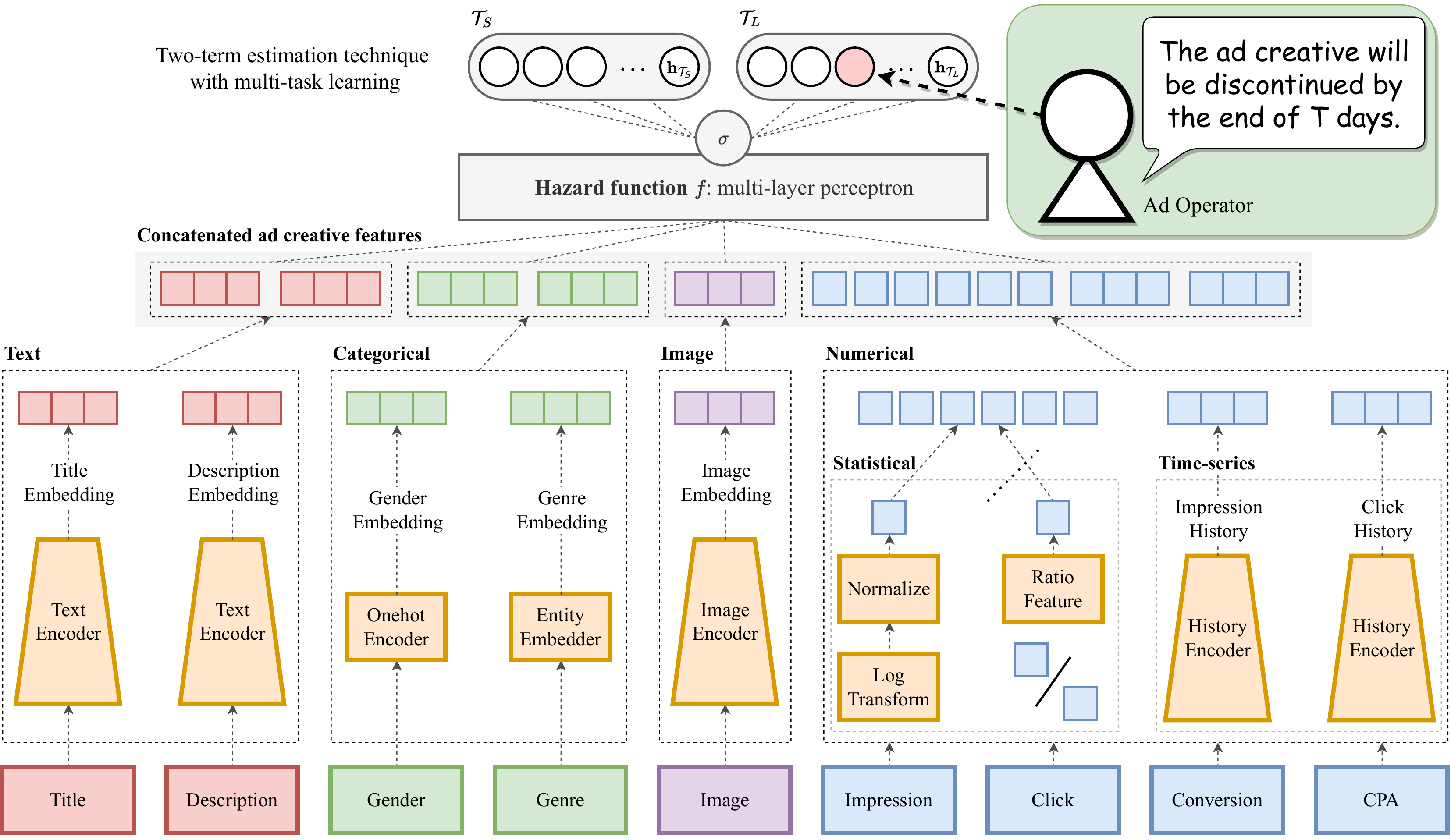}
    \caption[Outline of our framework]{
        Outline of our framework that exploits a hazard function, which draws on the idea of survival prediction, to predict the discontinuation of ad creatives.
        The input includes the four types of features: text, categorical, image, and numerical features.
        The output is the hazard probability, which includes whether the target ad creative has been discontinued in each time interval.
    }
    \label{fig:kitada2022ad/proposed_framework}
\end{figure*}

\subsection{Operational Support for Less Effective Ad}

Discontinuing ad creatives at an appropriate time is one of the most important ad operations that can have a significant impact on sales. 
Such operational support for ineffective ads has been less explored than that for effective ads. 
After pre-analyzing 1,000,000 real-world ad creatives, we found that there are two types of discontinuation: short-term (i.e., cut-out) and long-term (i.e., wear-out).
We propose a practical prediction framework for the discontinuation of ad creatives with a hazard function-based loss function inspired by survival prediction, as shown in Fig.~\ref{fig:kitada2022ad/proposed_framework}. 
Our framework accurately predicts the appropriate timing for the discontinuation of two types of digital advertisements, short-term and long-term. 
The framework consists of two main techniques: (1) a two-term estimation technique with multi-task learning and (2) a click-through rate-weighting technique for the loss function. 
We evaluated our framework using 1,000,000 real-world ad creatives, including 10 billion scale impressions. 

The attention mechanism incorporated in the proposed framework allows us to interpret the word expressions that contribute to ad discontinuation.
While various factors determine ad discontinuation, we expect that short- and long-term discontinuation in the ad text will show different trends. 
Specifically, in the short-term discontinuation, it is possible to identify word expressions in which the user did not show interest.
On the other hand, in the long-term discontinuation, it is possible to confirm word expressions that are no longer in season.
Based on the interpretation of these word expressions, the operator can make a final decision to discontinue the ad creatives.
Unfortunately, due to restrictions on information disclosure, quantitative evaluation of interpretability by specific attention visualization is not possible.
Meanwhile, we report that the performance of the framework is sufficient to support ad operators, and the word-by-word visualization the framework provides has significant advantages that could support discontinuation.

\section{Conclusion}
This summary provided an overview of the author's dissertation.
The dissertation discussed both basic and applied research on how attention mechanisms improve the performance and interpretability of machine learning models, especially in NLP-related tasks.
Focusing on the black-box nature of deep learning models, which have recently achieved significant results in various fields, we attempted to develop new training techniques and models that help to interpret the prediction results for the input words.
For the further development of applications of deep learning models, this black box nature is a serious barrier to analyzing the behavior of the models and using them in situations where prediction failures are not tolerated.
To the best of our knowledge, where there is a trade-off between prediction performance and interpretability, we have pioneered a new technique that aims to improve it.

Although the proposals within the dissertation are mainly validated using RNN models, we believe that the proposed methods are generic and can be applied to all models with attention mechanisms that will emerge in the future.
In interpreting the prediction evidence for each input word, we discussed the application of the proposals which entail a certain degree of interpretability.
Our ideas provide a positive effect on those that follow, and inform the emergence of further basic research into the development of the ideas, as well as efforts to support real-world operations.

Because there are several other options for providing explanations to DNNs besides attention mechanisms, we will confirm the applicability of the ideas in this method to these options in the future. 
Additionally, we will discuss the interpretability of the models in business applications, such as recommendation systems, to see if our method can be applied to such systems.

% or


% \appendices

% \section{}

\ifCLASSOPTIONcompsoc
  \section*{Acknowledgments}
\else
  \section*{Acknowledgment}
\fi

I would like to express my gratitude to my supervisor, Hitoshi Iyatomi. 
This work would not have been possible without him.
This work was supported by JSPS KAKENHI under Grant 21J14143.

\ifCLASSOPTIONcaptionsoff
  \newpage
\fi



%\bibliographystyle{IEEEtran}

% H.~Kopka and P.~W. Daly, \emph{A Guide to {\LaTeX}}, 3rd~ed.\hskip 1em plus

% \end{thebibliography}

\bibliographystyle{IEEEtran}
\bibliography{references}


% \end{IEEEbiography}


%\newpage



%\vfill


\end{document}